\documentclass[10pt,twocolumn,letterpaper]{article}

\usepackage[final]{cvpr}      

\definecolor{cvprblue}{rgb}{0.21,0.49,0.74}
\usepackage[pagebackref,breaklinks,colorlinks,allcolors=cvprblue]{hyperref}

\usepackage{amsmath}
\usepackage{amssymb}
\usepackage{booktabs}
\usepackage{xcolor}
\definecolor{cvprred}{rgb}{0.8, 0.1, 0.1} 

\usepackage[normalem]{ulem}

\usepackage[table]{xcolor}
\definecolor{highlightgray}{gray}{0.93}
\definecolor{LightBlue}{rgb}{0.88, 0.92, 1.0}
\definecolor{darkgreen}{rgb}{0.0, 0.6, 0.2}
\newcommand{\gain}[1]{\raisebox{-1.5pt}{\scriptsize\textcolor{darkgreen}{{$\uparrow$}#1}}}
\definecolor{darkred}{rgb}{0.8, 0.1, 0.1}

\usepackage{amsthm}

\usepackage[ruled,linesnumbered]{algorithm2e}
\usepackage{algorithm2e}
\SetKwInOut{KwIn}{Require}
\SetKwInOut{KwOut}{Ensure}

\usepackage{booktabs}
\usepackage{tabularx}

\def\paperID{} 
\def\confName{CVPR}
\def\confYear{2026}

\title{ConInfer: Context-Aware Inference for Training-Free Open-Vocabulary Remote Sensing Segmentation}

\author{
Wenyang Chen$^{1}$, 
Zhanxuan Hu$^{1}$, 
Yaping Zhang$^{1}$, 
Hailong Ning$^{2}$, 
Yonghang Tai$^{1}$ \thanks{Corresponding author} \\
$^{1}$Yunnan Normal University \\
$^{2}$Xi'an University of Posts \& Telecommunications \\
{\tt\small \{dog.yang000, zhanxuanhu\}@gmail.com, zhangyp@ynnu.edu.cn} \\
{\tt\small ninghailong@xupt.edu.cn, taiyonghang@126.com}
}

\begin{document}
\maketitle
\begin{abstract}
Training-free open-vocabulary remote sensing segmentation (OVRSS), empowered by vision-language models, has emerged as a promising paradigm for achieving category-agnostic semantic understanding in remote sensing imagery. Existing approaches mainly focus on enhancing feature representations or mitigating modality discrepancies to improve patch-level prediction accuracy. However, such independent prediction schemes are fundamentally misaligned with the intrinsic characteristics of remote sensing data. In real-world applications, remote sensing scenes are typically large-scale and exhibit strong spatial as well as semantic correlations, making isolated patch-wise predictions insufficient for accurate segmentation.
To address this limitation, we propose \textbf{ConInfer}, a context-aware inference framework for OVRSS that performs joint prediction across multiple spatial units while explicitly modeling their inter-unit semantic dependencies.
By incorporating global contextual cues, our method significantly enhances segmentation consistency, robustness, and generalization in complex remote sensing environments. Extensive experiments on multiple benchmark datasets demonstrate that our approach consistently surpasses state-of-the-art per-pixel VLM-based baselines such as SegEarth-OV, achieving average improvements of 2.80\% and 6.13\% on open-vocabulary semantic segmentation and object extraction tasks, respectively. The implementation code is available at: \url{https://github.com/Dog-Yang/ConInfer}
\end{abstract}
    
\section{Introduction}
\label{sec:intro}

\begin{figure}[t]
    \centering
    \includegraphics[width=0.8\linewidth]{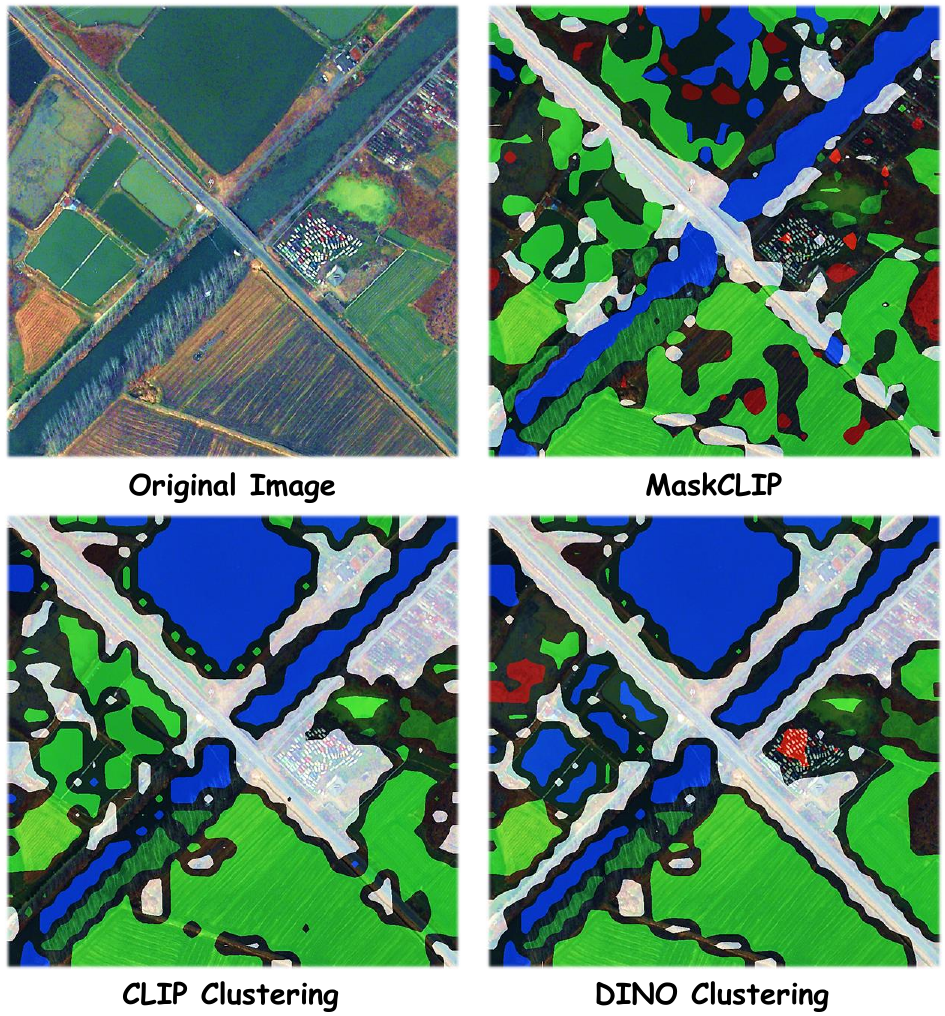}
    \caption{
    Compared with the direct predictions of MaskCLIP, the segmentation maps generated by clustering patch features from CLIP or DINOv3 exhibit superior spatial consistency.
    }
    \label{fig:intro}
\end{figure}

Semantic segmentation plays a crucial role in remote sensing by enabling detailed scene understanding and supporting diverse downstream applications such as land cover mapping~\cite{land_cover}, urban planning~\cite{building_seg}, and environmental monitoring~\cite{environmental_monitoring}. Traditional remote sensing segmentation methods~\cite{MSGCNet,UNetFormer,CADANet}, however, are typically developed under \textit{closed-set assumptions}, where the model is restricted to a fixed set of predefined categories. Such a setting fails to reflect dynamic real-world scenarios, where novel or unseen objects frequently appear in large-scale and heterogeneous environments. This limitation motivates the exploration of \textit{Open-Vocabulary Remote Sensing Segmentation}~\cite{GSNet,Segearth-ov,RSVG-ZeroOV}, which aims to achieve category-agnostic semantic understanding and generalize to arbitrary concepts beyond the training label space. 

Empowered by the zero-shot capabilities of vision-language models (VLMs), train-free open-vocabulary segmentation~(OVS) has achieved remarkable success on natural images, and most existing train-free OVRSS approaches directly build upon these techniques to process remote sensing data. Two major research directions have emerged: (1) enhancing patch-level representations~\cite{Segearth-ov,CorrCLIP} to strengthen the discriminative power of visual tokens for dense prediction, and (2) mitigating the modality gap between visual and textual spaces~\cite{T2I_Activation,Talking2DINO} to improve cross-modal consistency. While these methods have achieved promising results, they largely inherit the paradigm of \textit{independent prediction}, where each patch is treated as an isolated unit.

However, remote sensing imagery differs fundamentally from natural images. Remote sensing scenes are captured at large spatial scales, encompassing heterogeneous objects (\eg buildings, vegetation, water bodies) with complex spatial organization and strong semantic correlations~\cite{rolf2024mission}. Moreover, visually similar regions (\eg roads and runways) may convey different semantics depending on their spatial context, making purely independent predictions prone to fragmentation or misclassification. Consequently, directly transferring OVS techniques from natural images fails to fully exploit the contextual dependencies intrinsic to remote sensing imagery, leading to suboptimal segmentation consistency in large-scale scenes, as shown in~\cref{fig:intro}.

To overcome these limitations, we argue that contextual information should be explicitly incorporated during inference to capture spatial dependencies and improve segmentation consistency. As illustrated in~\cref{fig:intro}, compared to the results obtained from a text classifier, clustering the patch features extracted by DINOv3~\cite{DINOv3} produces segmentation maps with superior spatial coherence. However, DINOv3 features alone lack semantic alignment with the language space, making them unsuitable for direct open-vocabulary prediction. 
To this end, we propose a \textit{Context-aware Inference framework~(\textbf{ConInfer})} for OVRSS that jointly exploits the strengths of DINOv3 and VLMs. 
Specifically, our approach leverages the contextual cues encoded by DINOv3 to refine the VLM’s predictions, enforcing spatial and semantic consistency across patches~(See~\cref{fig:frame}). This design effectively bridges the gap between semantic understanding and spatial reasoning. 
Notably, our framework is entirely \textit{training-free} and orthogonal to existing methods, allowing it to serve as a \textit{plug-and-play module} for enhancing various VLM-based segmentation models. In comparison with Segearth-OV~\cite{Segearth-ov}, our approach build upon a simple baseline~\cite{ClearCLIP} outperforms it by 2.80\% and 6.13\% mIOU on open-vocabulary semantic segmentation, and object extraction tasks, respectively. The main contributions of this work can be summarized as follows:

\begin{itemize}
    \item \textbf{A new perspective.} To the best of our knowledge, this is the \textit{first work} to explicitly incorporate contextual information of remote sensing imagery during the inference stage, offering a new perspective on open-vocabulary remote sensing segmentation.
    
    \item \textbf{A simple yet effective framework.} We propose \textit{a simple yet effective context-aware inference framework} that establishes a paradigm for integrating the contextual reasoning capability of vision foundation models with the semantic alignment ability of vision-language models.
    
    \item \textbf{Promising results.} Despite its simplicity and training-free nature, our approach achieves consistently superior performance across multiple benchmarks, showing clear potential for advancing OVRSS in real-world scenarios.
\end{itemize}

\section{Related Work}
\label{sec:related}

\subsection{Open-Vocabulary Segmentation}
\noindent
OVS aims to enable pixel-level recognition for arbitrary categories beyond the closed-set vocabulary, typically by leveraging the zero-shot capabilities of VLMs~\cite{transclip}. Existing studies can be broadly categorized into \textit{training-based}~\cite{SED,SAN,DeCLIP,GroupViT,CLIP-DINOiser} and \textit{training-free}~\cite{MaskCLIP,ClearCLIP,CorrCLIP,Proxyclip,ReME} approaches. The former often fine-tune the visual or textual components of VLMs with additional supervision, while the latter perform segmentation directly using additional pre-trained models, such as DINO~\cite{DINOv2,DINOv3} and SAM~\cite{SAM,SAM2}, without any further training. Compared with training-based approaches, training-free methods avoid the additional cost of annotation and large-scale computation, making them increasingly attractive to the research community.

In the natural image domain, training-free OVS has achieved remarkable progress by adapting image-level representations from VLMs for dense, pixel-level prediction. Early studies focused on enhancing the spatial localization of visual features, for instance, by refining the attention maps or activations from VLMs to improve object boundary delineation~\cite{MaskCLIP,SCLIP,NACLIP,ResCLIP,ClearCLIP}. Subsequent works sought to enrich representation quality and reduce the modality gap through multi-scale feature fusion~\cite{GEM,SC-CLIP}or prototype optimization~\cite{FreeDA,OVDiff,ReME}. Inspired by these advances, recent efforts have explored extending OVS frameworks to remote sensing imagery~\cite{Segearth-ov,GSNet,RSKT-Seg,SCORE}. Typical strategies include domain-specific feature calibration~\cite{RSKT-Seg,Segearth-ov}, prototype alignment~\cite{Segclip,TPOV-Seg}, or pseudo-label-based supervision to mitigate the domain gap between natural and remote sensing images~\cite{T2I_Activation}. While such methods show promising results, they mostly inherit the paradigm of \textit{independent prediction}, neglecting the strong spatial and semantic correlations~\cite{rolf2024mission} of remote sensing data. Consequently, \textit{independent prediction} fails to fully exploit the contextual dependencies intrinsic to remote sensing imagery.

\begin{figure*}[t]
    \centering
    \includegraphics[width=1.0\textwidth]{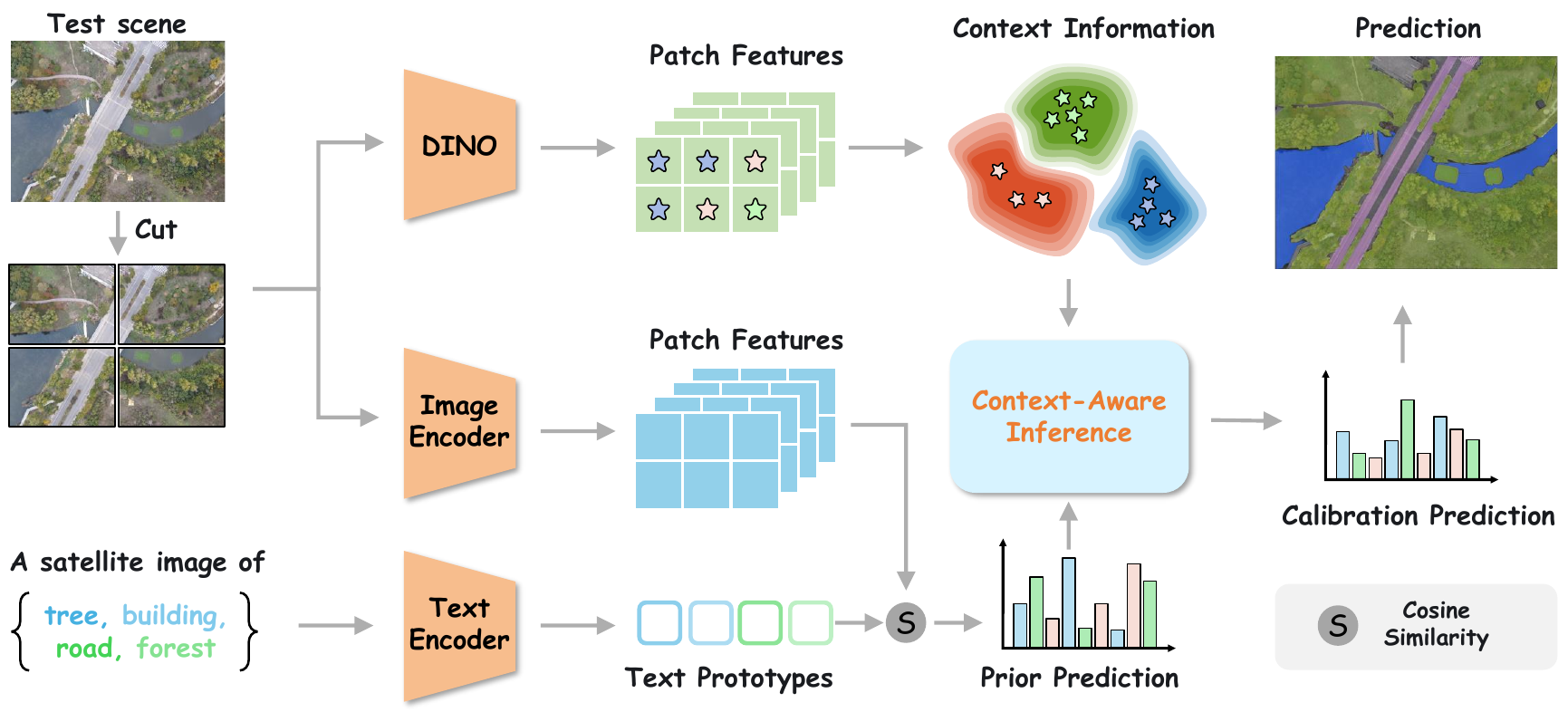}
    \caption{Overview of the proposed Context-Aware Inference~(\emph{ConInfer}) framework. \emph{ConInfer} performs joint, training-free inference by integrating semantic priors from CLIP with structural regularization from DINO. 
    This formulation enables globally consistent open-vocabulary segmentation without any retraining.}
    \label{fig:frame}
\end{figure*}
\subsection{Context Modeling in Segmentation}
Context modeling forms a cornerstone in semantic segmentation, aiming to capture spatial and semantic dependencies among pixels or regions to resolve ambiguities arising from local predictions and ensure global consistency of segmentation results. Existing methods focus on aggregating contextual information via multi-scale feature fusion~\cite{OVRS} and feature aggregation techniques~\cite{SCORE}. In the remote sensing domain, due to the large spatial scale, dense distribution of objects, and strong spatial correlations, the proper exploitation of contextual cues is essential for suppressing fragmented predictions and improving the coherence of overall scene structure~\cite{rolf2024mission}.

However, a common characteristic of current methods is that they integrate context modeling into the training stage by embedding attention modules, graph convolutional layers, or multi-scale fusion units within the network architecture~\cite{GeoRSCLIP,RemoteCLIP,SkyCLIP}. This paradigm suffers from two main limitations: first, the context modules are tightly coupled with backbone networks, reducing flexibility and generality; second, they cannot operate independently during inference, thus failing to serve as plug-and-play components for arbitrary pre-trained VLMs. In training-free OVRSS, such structural dependencies constrain the generalization of pre-trained VLMs and weaken their adaptability to novel categories and large-scale scenes.

\section{Method}
\label{sec:Method}
\cref{fig:frame} illustrates the overall pipeline of our proposed \emph{ConInfer} framework. 
{Given a large-scale remote sensing scene}, we first partition it into fixed-size, non-overlapping tiles. {Each tile is processed independently} to extract two types of patch-level features: (1) high-level contextual features from a vision foundation model (\eg DINOv3) that capture structural patterns within the tile, and (2) semantic-aligned visual features from \emph{any} VLM-based OVRSS model that produce open-vocabulary recognition scores via cosine similarity with text prototypes.
The DINO-derived features implicitly encode \emph{contextual information} across patches, enabling spatial grouping of visually coherent regions that often correspond to semantically related areas in remote sensing imagery. Meanwhile, the VLM branch yields a \emph{prior prediction} distribution for each patch from text prototypes, but operates independently and is agnostic to scene context. \emph{ConInfer} bridges these two complementary sources through a \emph{training-free context-aware inference} module, yielding a \emph{calibrated prediction} that integrates semantic alignment from the language domain with contextual reasoning from the visual domain. In the following sections, we present the design and formulation of each component within \emph{ConInfer}.

\subsection{Prior Prediction from VLM-based OVRSS}

We follow the standard OVRSS paradigm to generate the \emph{prior prediction}. Specifically, given an input tile $\mathcal{I}$, a vision encoder $f(\cdot)$ extracts patch-level features:
\begin{equation}
\mathbf{V} = f(\mathcal{I}) \in \mathbb{R}^{n \times d},
\end{equation}
where $n$ denotes the number of patches, and $d$ denotes the feature dimension. For a candidate category set $\mathcal{C}$, each category name $c \in \mathcal{C}$ is converted into a prompt template $T(c)$ and encoded using a text encoder $g(\cdot)$ to obtain text prototype vectors:
\begin{equation}
\mathbf{T} = \{ g(T(c)) \mid c \in \mathcal{C} \} \in \mathbb{R}^{|\mathcal{C}| \times d}.
\end{equation}
Prediction is then achieved by computing patch-to-text similarity via cosine similarity:
\begin{equation}
\boldsymbol s_{ij} = \mathrm{cos}(\mathbf{V}_i , \mathbf{T}_j).
\end{equation}
After softmax normalization, we obtain the per-patch category probability distribution:
\begin{equation}
\boldsymbol p_{ij} = \frac{\exp(\boldsymbol s_{ij}/\tau)}{\sum_{k=1}^{|\mathcal{C}|} \exp(\boldsymbol s_{ik}/\tau)}.
\end{equation}
The resulting distribution $p_{ij}$ forms the \emph{prior prediction} for each patch. Importantly, the VLM-based OVRSS model used here is not restricted to a specific architecture or implementation: any existing OVRSS model can serve as this branch. \emph{ConInfer} only consumes prediction scores and does not modify or fine-tune any model parameters.

\subsection{Context Information Extraction from VFMs}

To capture contextual dependencies in remote sensing imagery, we leverage a vision foundation model (VFM), specifically DINOv3, to extract high-level visual features from all patches across all tiles of the scene. Unlike the VLM-based OVRSS branch, which is aligned with textual semantics, the VFM is trained purely on visual data and therefore encodes structural layout, spatial organization, and appearance regularities present within the scene.

Let $\mathbf{X} = \{\mathbf{x}_1, \mathbf{x}_2, \ldots, \mathbf{x}_N\}$ denote the set of DINO-derived patch features, where each $\mathbf{x}_i \in \mathbb{R}^{d}$ is one patch feature in the entire scene. To model the contextual structure, we fit a Gaussian Mixture Model (GMM) with $|\mathcal{C}|$ mixture components:
\begin{equation}
p(\mathbf{x}) = \sum_{k=1}^{|\mathcal{C}|} \boldsymbol \pi_k \mathcal{N}(\mathbf{x} \mid \boldsymbol \mu_k, \boldsymbol \Sigma_k),
\end{equation}
where $\pi_k$, $\mu_k$, and $\Sigma_k$ denote the mixture weight, mean, and covariance of each Gaussian component, respectively. The GMM parameters can be estimated using Expectation-Maximization (EM) in a training-free manner.

For each patch feature $\mathbf{x}_i$, we compute its posterior assignment over mixture components:
\begin{equation}
\mathbf{q}_i = [q_{i1}, \ldots, q_{i|\mathcal{C}|}], \quad q_{ik} = p(z=k \mid \mathbf{x}_i),
\end{equation}
which reveals the context information. Patches belonging to semantically relevant or spatially coherent regions tend to exhibit similar posterior distributions $\mathbf{q}_i$. As such, the GMM posterior encapsulates structural context cues that reflect \emph{how patches are spatially and visually related across the scene}.
However, despite capturing strong contextual regularities, the GMM posterior does not reveal semantic meaning because it is not aligned with the textual label space. This complementary property naturally motivates the next step.
\subsection{Context-Aware Inference}

Given the semantic prior distributions $\{\mathbf{p}_i\}_{i=1}^N$ (from the OVRSS branch) and the contextual posterior distributions $\{\mathbf{q}_i\}_{i=1}^N$ (from the VFM branch), our goal is to obtain calibrated probability distributions $\{\mathbf{z}_i\}_{i=1}^N$. The calibrated distributions should jointly reflect both semantic alignment (from VLM) and contextual coherence (from VFM).

\paragraph{Joint Latent Distribution Optimization.}
For each patch, we formulate $\mathbf{z}_i$ as a latent categorical distribution that is encouraged to stay close to both sources of information. We achieve this by minimizing the sum of two KL divergences:
\begin{equation}\label{eq:kl-basic}
\min_{\{\mathbf{z}_i\}_{i=1}^{N}} \; \sum_{i=1}^N \Big[\, \mathrm{KL}\!\left(\mathbf{z}_i \,\|\, \mathbf{p}_i\right) + \mathrm{KL}\!\left(\mathbf{z}_i \,\|\, \mathbf{q}_i\right) \,\Big],
\end{equation}
where $N$ denotes the number of patches. Eq.~\eqref{eq:kl-basic} encourages $\mathbf{z}_i$ to lie in the agreement region between the semantic prediction $\mathbf{p}_i$ and the contextual structure encoded by $\mathbf{q}_i$.

Directly optimizing Eq.~\eqref{eq:kl-basic} is of limited value because $\mathbf{q}_i$ is obtained from a context model learned \emph{without} semantic information; thus $\mathbf{q}_i$ is not aligned with $\mathbf{p}_i$ in the semantic space.
A straightforward remedy is to initialize the GMM with the semantic prior $\mathbf{p}_i$ so that the learned $\mathbf{q}_i$ carries initial semantic alignment; one can then apply Eq.~\eqref{eq:kl-basic} to obtain a consensus prediction~\cite{GDA}.
However, such an isolated procedure prevents the GMM parameter estimation from benefiting from the evolving $\mathbf{z}_i$. We therefore propose \emph{joint learning} of the latent consensus distributions and the GMM parameters, leading to the following objective:
\begin{equation}\label{eq:kl-joint}
\min_{\{\mathbf{z}_i\}_{i=1}^{N},\,\mathbf{\mu},\,\mathbf{\Sigma}} \; \sum_{i=1}^N \Big[\, \mathrm{KL}\!\left(\mathbf{z}_i \,\|\,\mathbf{p}_i\right) + \mathrm{KL}\!\left(\mathbf{z}_i \,\|\, \mathbf{q}_i\right) \,\Big],
\end{equation}
where $\mathbf{q}_i$ is the GMM posterior computed from the DINO-derived features and depends on the GMM parameters (means $\mathbf{\mu}$ and covariances $\mathbf{\Sigma}$).
\footnote{The mixture weights $\pi$ are held fixed to a uniform distribution ($\pi_k = 1/|\mathcal{C}|$) and are therefore not optimization variables.} 
This joint objective enables $\mathbf{z}_i$ and $\mathbf{q}_i$ to be mutually refined: $\mathbf{z}_i$ becomes increasingly context-aware while $\mathbf{q}_i$ gradually absorbs semantic alignment from the VLM prior $\mathbf{p}_i$. Ablation results on joint learning are shown in~\cref{tab:ablation_study}.

\subsection{Optimization}
We optimize Eq.~\eqref{eq:kl-joint} using an alternating manner.

\vspace{2pt}\noindent\textbf{Initialization.}
We first initialize the GMM parameters using the semantic prior distribution $\boldsymbol p_i$. Let $\boldsymbol q_{ik}^{(0)} \propto \boldsymbol p_{ik}$ denote the initial responsibilities, based on which initial mixture parameters are computed as:
\begin{align}
\boldsymbol \mu_k^{(0)} &= \frac{\sum_{i} \boldsymbol q_{ik}^{(0)}\,\mathbf{x}_i}{\sum_{i} \boldsymbol q_{ik}^{(0)}}, \label{eq:mu_init}\\
\boldsymbol \Sigma_k^{(0)} &= \frac{\sum_{i} \boldsymbol q_{ik}^{(0)}(\mathbf{x}_i- \boldsymbol \mu_k^{(0)})(\mathbf{x}_i- \boldsymbol \mu_k^{(0)})^\top}{\sum_{i} \boldsymbol q_{ik}^{(0)}}. \label{eq:sigma_init}
\end{align}

\paragraph{Alternating Updates.}
Given the initialization, we iteratively update $\mathbf{q}$, $\mathbf{z}$, and the GMM parameters. For notational simplicity, we omit the superscripts hereafter.

\textbf{(1) Update $\boldsymbol q$ (E-step).} With the current mixture parameters $\{\pi_k,\mu_k,\Sigma_k\}$, the GMM posterior is obtained by:
\begin{equation}
q_{ik}=\frac{\boldsymbol \pi_k\,\mathcal{N}(\mathbf{x}_i\mid \boldsymbol \mu_k, \boldsymbol \Sigma_k)}{\sum_{\ell=1}^{C}\pi_\ell\,\mathcal{N}(\mathbf{x}_i\mid \boldsymbol \mu_\ell, \boldsymbol \Sigma_\ell)}.
\label{eq:compute_q}
\end{equation}

\textbf{(2) Update $\boldsymbol z$.} Given the semantic prior $\mathbf{p}_i$ and the contextual posterior $\mathbf{q}_i$, the closed-form solution of $\mathbf{z}$ minimizing the KL-based agreement objective is:
\begin{equation}
\boldsymbol z_{ik} \propto {\boldsymbol p_{ik}\, \odot  \boldsymbol q_{ik}}, \qquad \sum_{k=1}^{C} \boldsymbol z_{ik}=1.
\label{eq:update_z}
\end{equation}

\textbf{(3) Update GMM (M-step).} We treat the updated $\mathbf{z}_i$ as soft responsibilities and re-estimate mixture parameters:
\begin{align}
\boldsymbol \mu_k &= \frac{\sum_{i} \boldsymbol z_{ik}\,\mathbf{x}_i}{\sum_{i} \boldsymbol z_{ik}}, \label{eq:mu_update} \\
\boldsymbol \Sigma_k &= \frac{\sum_{i} \boldsymbol z_{ik} (\mathbf{x}_i- \boldsymbol \mu_k)(\mathbf{x}_i- \boldsymbol \mu_k)^\top}{\sum_{i} \boldsymbol z_{ik}}. \label{eq:sigma_update}
\end{align}

In practice, these three steps are executed in an alternating manner until the algorithm converges, which usually takes fewer than 10 iterations (See~\cref{fig:Loss}).

\subsection{Convergence Analysis}
Our alternating optimization process shows stable convergence behavior in practice. In each iteration, updating $\boldsymbol z$ decreases the objective by moving it toward the agreement region of $\boldsymbol p$ and $\boldsymbol q$, while updating $\boldsymbol q$ (via GMM re-estimation) makes the contextual posterior better match the consensus distribution $\boldsymbol z$. As a result, the two steps monotonically reduce the objective value and progressively align $\boldsymbol z$ and $\boldsymbol q$. We empirically observe that the optimization converges within a small number of iterations (See~\cref{fig:Loss}.)

\subsection{Final Prediction}
At inference time, we directly use the calibration prediction $\boldsymbol Z$ as the semantic probability for each patch. Since prediction is performed at the patch level, the final segmentation mask for the large-scale remote sensing scene is obtained by mapping each patch back to its original tile position and performing tile-wise upsampling to reconstruct full-resolution dense segmentation.

\begin{algorithm}[t]
\caption{Optimization}
\label{alg:ConInfer}
\KwIn{VFMs features $\mathbf X \in \mathbb{R}^{N \times d}$, VLMs prior prediction $\mathbf P \in \mathbb{R}^{N \times C}$, max iterations $L$}
\KwOut{Consensus distribution $\mathbf Z$ and GMM parameters $(\boldsymbol \mu, \boldsymbol \Sigma)$}

Initialize GMM parameters $(\boldsymbol \mu^{0}, \boldsymbol \Sigma^{0})$ using $\mathbf P$ via~\cref{eq:mu_init,eq:sigma_init}; \\

\For{$l = 1$ to $L$}{
    Compute contextual posteriors $\mathbf Q$ via GMM E-step using ~\cref{eq:compute_q}; \\

    Update consensus distribution $\mathbf Z$ via~\cref{eq:update_z}; \\

    Re-estimate $(\boldsymbol \mu, \boldsymbol \Sigma)$ via GMM M-step using $\mathbf Z$ according to~\cref{eq:mu_update,eq:sigma_update}; \\
}

\Return{$\mathbf Z, (\boldsymbol \mu, \boldsymbol \Sigma)$}
\end{algorithm}
\section{Experiments}
\label{sec:experiments}

\begin{table*}[t]
  \centering
  \caption{Quantitative comparison of open-vocabulary semantic segmentation performance (mIoU, \%) across eight remote sensing datasets. These datasets comprise high-resolution satellite imagery (first five) and UAV-based urban scenes (latter three). \textbf{Bold} and \underline{underlined} values indicate the best and second-best performance, respectively.}
  \label{tab:multi_seg_comparison}
  \resizebox{\textwidth}{!}{%
  \begin{tabular}{lccccccccc}
  \toprule
 & \multicolumn{5}{c}{Satellite Images}& \multicolumn{3}{c}{UAV Images}&\\
 \cmidrule(lr){2-6} \cmidrule(lr){7-9}
    \textbf{Methods} & \textbf{OpenEarthMap} & \textbf{LoveDA} & \textbf{iSAID} & \textbf{Potsdam} & \textbf{Vaihingen} & \textbf{UAVid} & \textbf{UDD5} & \textbf{VDD} & \textbf{Average} \\
    \midrule
    CLIP~\cite{CLIP} & 9.16 & 9.59 & 6.53 & 11.62 & 7.84 & 8.94 & 7.91 & 11.27 & 9.11 \\
    MaskCLIP~\cite{MaskCLIP} & 22.30 & 25.25 & 14.32 & 31.41 & 22.78 & 30.09 & 32.74 & 29.12 & 26.13 \\
    SCLIP~\cite{SCLIP} & 23.85 & 24.60 & 15.08 & 31.84 & 21.96 & 29.69 & 33.25 & 34.26 & 26.82 \\
    GEM~\cite{GEM} & 29.96 & 26.95 & {\uline{20.01}} & 38.24 & 21.59 & 33.49 & 36.02 & 38.46 & 30.59 \\
    ClearCLIP~\cite{ClearCLIP} & 29.94 & 28.06 & 16.62 & 40.20 & 20.55 & 34.35 & 38.07 & 40.24 & 31.00 \\
    SegEarth-OV~\cite{Segearth-ov} & {\uline{37.81}} & {\uline{35.79}} & 18.14 & {\uline{47.94}} & {\uline{24.96}} & {\uline{42.81}}& {\textbf{49.64}} & {\uline{46.73}}& {\uline{37.98}} \\
    \rowcolor{LightBlue}
    MaskCLIP*(baseline) & 32.46 & 31.58 & 18.17 & 44.49 & 19.94 & 37.20 & 42.17 &  42.54& 33.57\\
    \rowcolor{LightBlue}
    ConInfer(Ours) & {\textbf{41.95}}\gain{9.49} & {\textbf{39.33}}\gain{7.75} & {\textbf{20.08}}\gain{1.91} & {\textbf{49.99}}\gain{5.50} & {\textbf{31.37}}\gain{11.43} & {\textbf{46.40}}\gain{9.20}& {\uline{46.86}}\gain{4.69} & {\textbf{50.29}}\gain{7.75}& {\textbf{40.78}}\gain{7.21}\\
    \bottomrule
  \end{tabular}%
  }
\end{table*}

\subsection{Dataset}
We follow the evaluation protocol of SegEarth-OV~\cite{Segearth-ov} and conduct experiments on 17 open-vocabulary remote sensing segmentation datasets, covering both multi-class semantic segmentation and single-class extraction tasks. 
For multi-class semantic segmentation, we evaluate on eight datasets: OpenEarthMap~\cite{Openearthmap}, LoveDA~\cite{LoveDA}, iSAID~\cite{iSAID}, Potsdam, Vaihingen\footnote{https://www.isprs.org/education/benchmarks/UrbanSemLab}, UAVid~\cite{UAVid}, UDD5~\cite{UDD}, and VDD~\cite{vdd}. 
These datasets encompass diverse land-cover categories and varied spatial structures, with the first five mainly containing high-resolution satellite imagery and the latter three focusing on unmanned aerial vehicle~(UAV) urban scene observations. 
For single-class extraction, we concentrate on three key tasks: building extraction, using four datasets (WHU$^{\text{Aerial}}$~\cite{WHU}, WHU$^{\text{Sat.II}}$~\cite{WHU}, Inria~\cite{Inria}, and xBD$^{\text{pre}}$~\cite{xBD}); road extraction, also with four datasets (CHN6-CUG~\cite{CHN6-CUG}, DeepGlobe\footnote{http://deepglobe.org}, Massachusetts~\cite{Massachusetts}, and SpaceNet~\cite{Spacenet}); and flood detection, with the WBS-SI5 dataset\footnote{https://www.kaggle.com/datasets/shirshmall/water-body-segmentation-in-satellite-images}.
These latter datasets emphasize fine-grained structural targets (\eg, buildings and roads) or dynamic hydrological regions. See Appendix for dataset details.

\begin{figure}[t]
    \centering
    \includegraphics[width=0.9\linewidth]{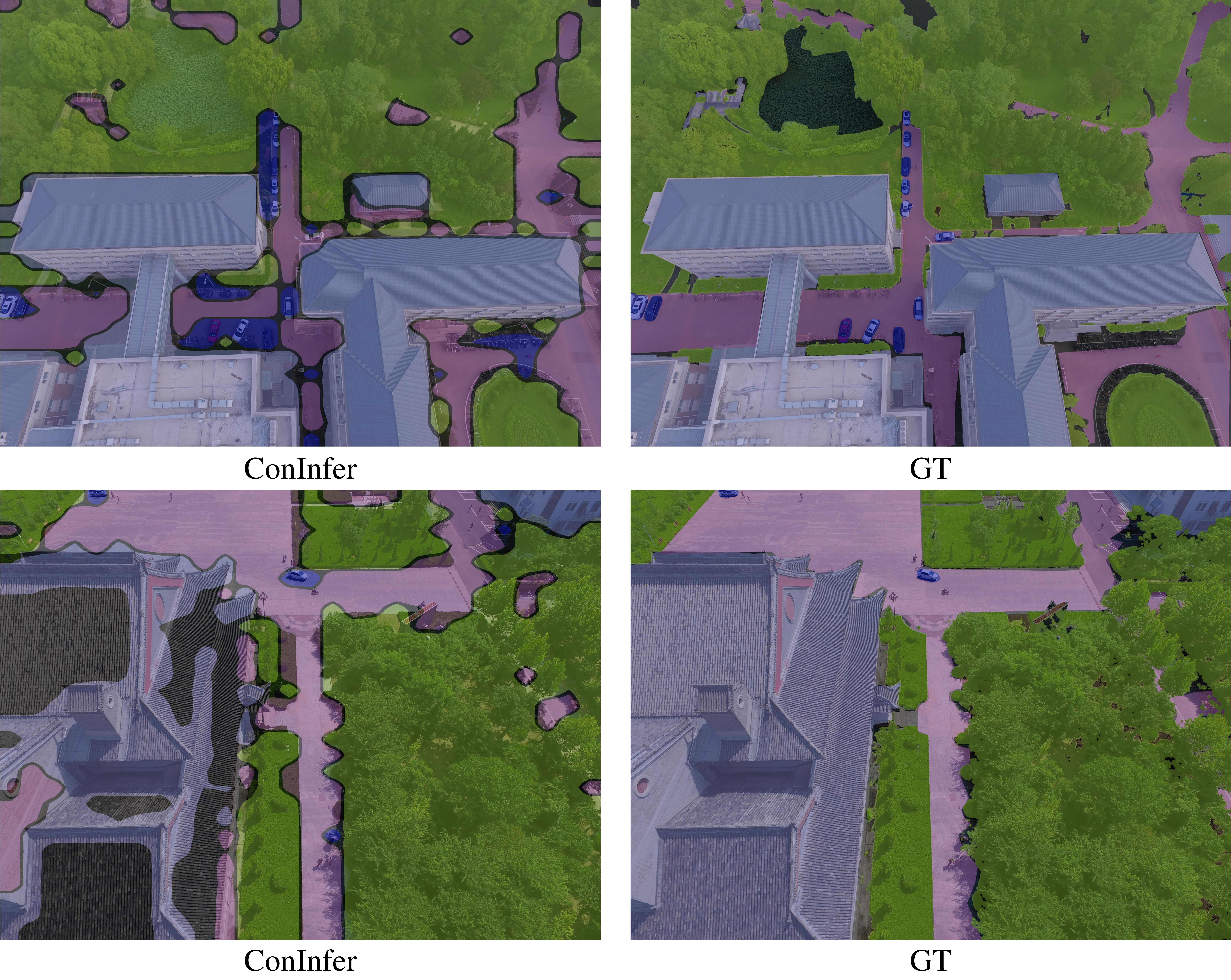}
    \caption{
    Examples from the UDD5 dataset illustrating cases where ConInfer encounters difficulties, including weak boundaries on small objects (vehicles) and category confusion (roof tiles).
    }
    \label{fig:udd}
\end{figure} 

\subsection{Setup}
\paragraph{Implementation.} 
Our implementation is built upon the MMSegmentation toolkit\footnote{https://github.com/open-mmlab/mmsegmentation}
. For the vision backbones, we load CLIP~\cite{CLIP} using the official ViT-B/16 pretrained weights from OpenAI. For DINOv3, we adopt the ViT-H+/16 distilled model for the VDD and UAVid datasets, while using the ViT-L/16 distilled checkpoint pretrained on SAT-493M~\cite{DINOv3} for all other datasets. For the text encoder, we use the standard ImageNet prompt template (\eg “a photo of a {class name}”). In single-class extraction tasks, we further enhance the background prompt by composing it with multiple fine-grained negative concepts (\eg forest, vegetation, bare land, river, paved road); detailed prompts are provided in the Appendix. Input images are resized to 448$\times$448 pixels, and inference is performed without sliding windows. All experiments are conducted on a single NVIDIA RTX 4090 GPU. During inference, after processing every 50 images, we apply the iterative correction procedure with the number of iterations set to 10.

\vspace{2pt}\noindent\textbf{Evaluation Protocol.} 
We evaluate the performance on semantic segmentation and single-class extraction using mean Intersection over Union (mIoU) and Intersection over Union (IoU) of the foreground class, respectively.

\vspace{2pt}\noindent\textbf{Baseline.} 
As in natural image tasks, the MaskCLIP~\cite{MaskCLIP} framework is also applicable to remote sensing scenarios. Based on this framework, we remove the feed-forward network (FFN) and residual connections in the final Transformer layer, following~\cite{ClearCLIP}. In addition, we replace the final self-attention module with a multi-level fusion modulated attention mechanism~\cite{ResCLIP,Swin_transformer}, yielding our baseline model, MaskCLIP*. Further architectural details are provided in Appendix.

\subsection{Comparison with the State-of-the-Art Methods}
To evaluate the effectiveness of the proposed \emph{ConInfer} framework, we compare it against 6 state-of-the-art training-free OVS methods. This includes five representative approaches originally developed for natural image domains, namely Vanilla CLIP~\cite{CLIP}, MaskCLIP~\cite{MaskCLIP}, SCLIP~\cite{SCLIP}, GEM~\cite{GEM}, and ClearCLIP~\cite{ClearCLIP}, as well as SegEarth-OV~\cite{Segearth-ov}, the current state-of-the-art in OVRSS. Notably, SegEarth-OV introduces a universal feature upsampler during inference to refine dense predictions and better capture the fine-grained characteristics of remote sensing imagery.

\begin{table}[t]
  \centering
  \caption{Proposed method is a plug-and-play module that can be seamlessly integrated into existing OVS methods.}
  \label{tab:plug_and_play}
  \resizebox{\linewidth}{!}{
  \begin{tabular}{lccc}
    \toprule
    Methods     & OpenEarthMap & WHU$^{Aerial}$ & WBS-SI \\
    \midrule
    MaskCLIP~\cite{MaskCLIP}    & 22.30        & 32.00       & 40.30  \\
    +SegEarth-OV      & 25.94        & 38.18       & 47.75  \\
    \rowcolor{highlightgray}
    +ConInfer       & 29.28 \gain{6.98} & 54.98 \gain{22.98} & 52.43 \gain{12.13} \\
    \midrule
    SCLIP~\cite{SCLIP}       & 23.85        & 37.89       & 42.91  \\
    +SegEarth-OV      & 27.98        & 44.61       & 48.51  \\
    \rowcolor{highlightgray}
    +ConInfer       & 33.52 \gain{9.67} & 59.37 \gain{21.48} & 49.38 \gain{6.47} \\
    \midrule
    ClearCLIP~\cite{ClearCLIP}   & 29.94        & 43.59       & 46.55  \\
    +SegEarth-OV      & 36.12        & 49.89       & 55.80  \\
    \rowcolor{highlightgray}
    +ConInfer       & 41.36 \gain{11.42}  & 56.38 \gain{12.79} & 56.08 \gain{9.53} \\
    \bottomrule
  \end{tabular}
  }
\end{table}

\begin{figure*}[t]
    \centering
    \includegraphics[width=1.0\linewidth]{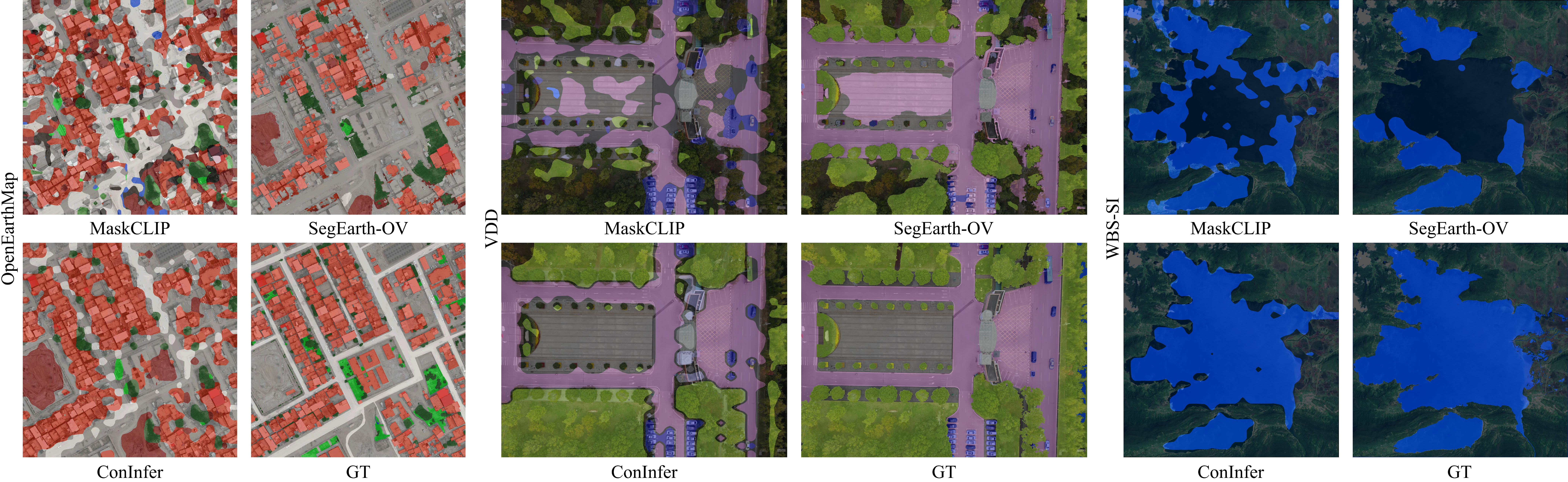}
    \caption{Qualitative comparison of different OVSS methods on the OpenEarthMap (land cover mapping), VDD (UAV aerial scene), and WBI-SI (water body segmentation) datasets. GT denotes the ground truth.}
    \label{fig:visual}
\end{figure*}

\begin{table*}[ht]
  \centering
  \caption{Quantitative comparison on the open-vocabulary single-class remote sensing extraction task, measured by \textit{Foreground} IoU. \textbf{Bold} and \underline{underlined} values indicate the best and second-best performance, respectively.}
  \label{tab:single_seg_comparison}
  \resizebox{\textwidth}{!}{%
  \begin{tabular}{lcccccccccc}\toprule
 & \multicolumn{4}{c}{Buiilding Extraction}& \multicolumn{4}{c}{Road Extraction}& Flood Detection&\\
    \cmidrule(lr){2-5} \cmidrule(lr){6-9} \cmidrule(lr){10-10} 
    \textbf{Method} & \textbf{WHU$^{Aerial}$} & \textbf{WHU$^{Sat.II}$} & \textbf{Inria} & \textbf{xBD$^{pre}$} & \textbf{CHN6-CUG} & \textbf{DeepGlobe} & \textbf{Massachusetts} & \textbf{SpaceNet} & \textbf{WBS-SI} & \textbf{Average} \\
    \midrule
    CLIP~\cite{CLIP} & 6.62 & 1.45 & 15.12 & 8.88 & 3.74& 1.13 & 3.50 & 3.20 & 8.98 & 5.85\\
    MaskCLIP~\cite{MaskCLIP} & 32.00 & 11.77 & 33.15 & 26.92 & 16.93 & 11.53 & 8.74 & 17.39 & 40.30 & 22.08 \\
    SCLIP~\cite{SCLIP} & 37.89 & 7.82 & 33.92 & 29.42 & 13.86 & 7.28 & 5.56& 8.13 & 42.91 & 20.75 \\
    GEM~\cite{GEM} & 41.48 & 15.98 & 40.74 & 33.48 & 22.40 & 13.43 & 7.51 & 17.92 & 50.64 & 27.06 \\
    ClearCLIP~\cite{ClearCLIP} & 43.59 & 13.72 & 37.08 & 29.83 & 25.20 & 12.86 & 7.18 & 17.02 & 46.55 & 25.89 \\
    SegEarth-OV~\cite{Segearth-ov} & \uline{52.90} & \uline{20.99} & \uline{45.41} & \uline{35.95} & {\uline{35.13}} & \uline{18.07} & \uline{10.43} & \uline{20.85} & \uline{57.38} & \uline{33.01} \\
    \rowcolor{LightBlue}
    MaskCLIP*(baseline) & 45.76 & 15.17 & 37.26 & 31.40 & 29.12 & 14.68 & 8.42 &  19.83 & 52.57 & 28.25 \\
    \rowcolor{LightBlue}
    ConInfer(Ours) & {\textbf{58.54}}\gain{12.78} & {\textbf{39.35}}\gain{24.18} & {\textbf{55.65}}\gain{18.39} & {\textbf{41.34}}\gain{9.94} & {\textbf{40.00}}\gain{10.88} & {\textbf{19.85}}\gain{5.17} & \textbf{12.16}\gain{3.74} & {\textbf{24.02}}\gain{4.19} & {\textbf{61.38}}\gain{8.81} & {\textbf{39.14}}\gain{10.89} \\
    \midrule
  \end{tabular}%
  }
\end{table*}

\vspace{2pt}\noindent\textbf{Semantic Segmentation.}
As shown in~\cref{tab:multi_seg_comparison}, the re-implemented MaskCLIP* baseline already brings notable gains over the original MaskCLIP, indicating that refining attention maps remains an effective means of enhancing training-free OVRSS performance. Building upon this stronger baseline, our \emph{ConInfer} further achieves consistent improvements across all eight datasets, reaching an average mIoU of 40.78\%. Compared to the strongest competitor, SegEarth-OV, \emph{ConInfer} attains a higher mean performance (+2.80\%) without relying on any pretrained upsampler, demonstrating that modeling cross-region semantic dependency at inference is a more direct and generalizable solution than introducing additional parametric modules. We note that \emph{ConInfer} shows slightly lower performance on the UDD5 dataset. Under low-altitude UAV viewpoints, the visual distribution differs substantially from satellite imagery, leading to two main issues when compared with the ground truth: (1) weak boundary sensitivity for small objects, such as vehicles, and (2) misclassification of roof tiles as background rather than belonging to the building class, as illustrated in~\cref{fig:udd}. Despite these issues, \emph{ConInfer} still produces coherent and semantically consistent segmentation results.

\vspace{2pt}\noindent\textbf{Single-Class Extraction.}
As presented in~\cref{tab:single_seg_comparison}, \emph{ConInfer} delivers notably larger gains in single-class extraction than in multi-class segmentation, increasing the baseline average IoU from 28.25\% to 39.14\% and surpassing SegEarth-OV by 6.13\% on average. 
The improvement is particularly significant on WHU$^{Sat.II}$, Inria, CHN6-CUG, and WBS-SI, indicating that context-aware inference is highly effective when the target class exhibits strong structural regularity. 
Buildings and roads typically form spatially continuous, homogeneous patterns, and the contextual priors encoded by DINOv3 allow \emph{ConInfer} to reinforce such structured dependencies, leading to more coherent and complete extraction regions. In contrast, independent prediction schemes often produce fragmented or discontinuous results. This trend is further corroborated by the visual comparisons in~\cref{fig:visual}, where \emph{ConInfer} yields segmentation masks with clearer boundaries and improved global connectivity.

\vspace{2pt}\noindent\textbf{Qualitative Results.}
\cref{fig:visual} presents a qualitative comparison of MaskCLIP, SegEarth-OV, and \emph{ConInfer}. Key observations include:
(1) MaskCLIP often produces fragmented regions and redundant classes (\eg bare ground misidentified as water) and suffers from notable misclassification (\eg rooftops labeled as background).
(2) SegEarth-OV provides finer details, but the lack of contextual reasoning leads to spatial discontinuities (\eg background appearing within buildings or missing regions in water bodies).
(3) \emph{ConInfer} delivers more coherent and semantically consistent segmentation by introducing context-aware inference. More visualizations can be found in Appendix.

\begin{table*}[ht]
  \centering
  \caption{
  Ablation study of \emph{ConInfer}. 
  We analyze the impact of alternative context information extraction (via CLIP features) and the removal of joint latent distribution optimization (applying prior knowledge post-EM).
  }
  \label{tab:ablation_study}
  \resizebox{\textwidth}{!}{%
  \begin{tabular}{l c c c c c c c c}
  \toprule
 & \multicolumn{4}{c}{Multi-class segmentation}& \multicolumn{4}{c}{Single-class extraction}\\
 \cmidrule(lr){2-5} \cmidrule(lr){6-9}
    \textbf{Methods} & \textbf{OpenEarthMap} & \textbf{LoveDA} & \textbf{iSAID} & \textbf{Vaihingen} & \textbf{WHU$^{Sat.II}$} & \textbf{Inria} & \textbf{xBD$^{pre}$} & \textbf{WBS-SI} \\
    \midrule
    Baseline & 32.46 & 31.58 & 18.17 & 19.94 & 15.17 & 37.26 & 31.40 & 52.57 \\
    Context$_\text{CLIP}$ & 35.12 \gain{2.66} & 33.21 \gain{1.63} & 19.05 \gain{0.88} & 25.10 \gain{5.16} & 17.38 \gain{2.21} & 38.27 \gain{1.01} & 34.98 \gain{3.58} & 54.48 \gain{1.91} \\
    No joint-learning & 36.23 \gain{3.77} & 36.27 \gain{4.69} & 18.86 \gain{0.69} & 32.80 \gain{12.86} & 28.56 \gain{13.39} & 54.47 \gain{17.21} & 36.30 \gain{4.90} & 56.78 \gain{4.21} \\
    ConInfer & {41.95} \gain{9.49} & {39.33} \gain{7.75} & {20.08} \gain{1.91} & {31.37} \gain{11.43} & {39.35} \gain{24.18} & {55.65} \gain{18.39} & {41.34} \gain{9.94} & {61.38} \gain{8.81} \\
    \bottomrule
  \end{tabular}%
  }
\end{table*}

\subsection{Ablation Studies}
\paragraph{\emph{ConInfer} is Plug-and-Play.}
By design, \emph{ConInfer} solely utilizes prediction scores and does not alter or fine-tune any model parameters, rendering it an ideal plug-and-play module for existing OVRSS frameworks. The results presented in~\cref{tab:plug_and_play} validate its exceptional efficacy in this capacity. When integrated, \emph{ConInfer} delivers significant and consistent performance gains to all baseline models (MaskCLIP, SCLIP, and ClearCLIP) across three distinct datasets. Notably, it boosts MaskCLIP's performance by an impressive 22.98\% on the WHU$^{Aerial}$ dataset. Furthermore, our approach consistently surpasses the performance of SegEarth-OV~\cite{Segearth-ov} across all experimental settings. These findings firmly establish \emph{ConInfer} as a highly effective, versatile, and superior solution for enhancing segmentation frameworks with minimal integration effort.

\vspace{2pt}\noindent\textbf{\emph{ConInfer} Benefits from Better VFMs.}
We investigate the impact of feature choice for context clustering by replacing DINOV3 features with visual features from CLIP. 
The results reproted in~\cref{tab:ablation_study} indicate consistent improvements across most datasets, with the exception of UDD5 and Potsdam, demonstrating the robustness of our overall design. 
However, the performance gains are marginal. 
We attribute this to CLIP's limited capability in capturing fine-grained details, which suggests that employing more powerful VFMs could lead to further enhancements.

\vspace{2pt}\noindent\textbf{\emph{ConInfer} Benefits from Joint-learning.}
To better understand the effect of the proposed joint optimization in Eq.~\eqref{eq:kl-joint}, we compare it with a \emph{Decoupled-learning} variant. The decoupled baseline follows a two-stage pipeline:  (1) the GMM is initialized using the VLM prior $\mathbf{p}_i$, and its parameters are then optimized independently on DINO features to obtain $\mathbf{q}_i$;  
(2) the final prediction is obtained by fusing the fixed GMM posterior $\mathbf{q}_i$ with the VLM prior $\mathbf{p}_i$ through Eq.~\eqref{eq:kl-basic}. As shown in~\cref{tab:ablation_study}, \emph{Decoupled-learning} can still yield noticeable improvements. However, its performance consistently lags behind \emph{Joint-learning}. The reason is that \emph{Joint-learning} establishes a bi-directional refinement loop: as $\mathbf{z}_i$ becomes more semantically meaningful, it reshapes the GMM’s posterior structure, which in turn provides more reliable contextual cues for updating $\mathbf{z}_i$. This iterative interplay enables the contextual model and semantic priors to reinforce each other.

\begin{figure}[t]
    \centering
    \includegraphics[width=1.0\linewidth]{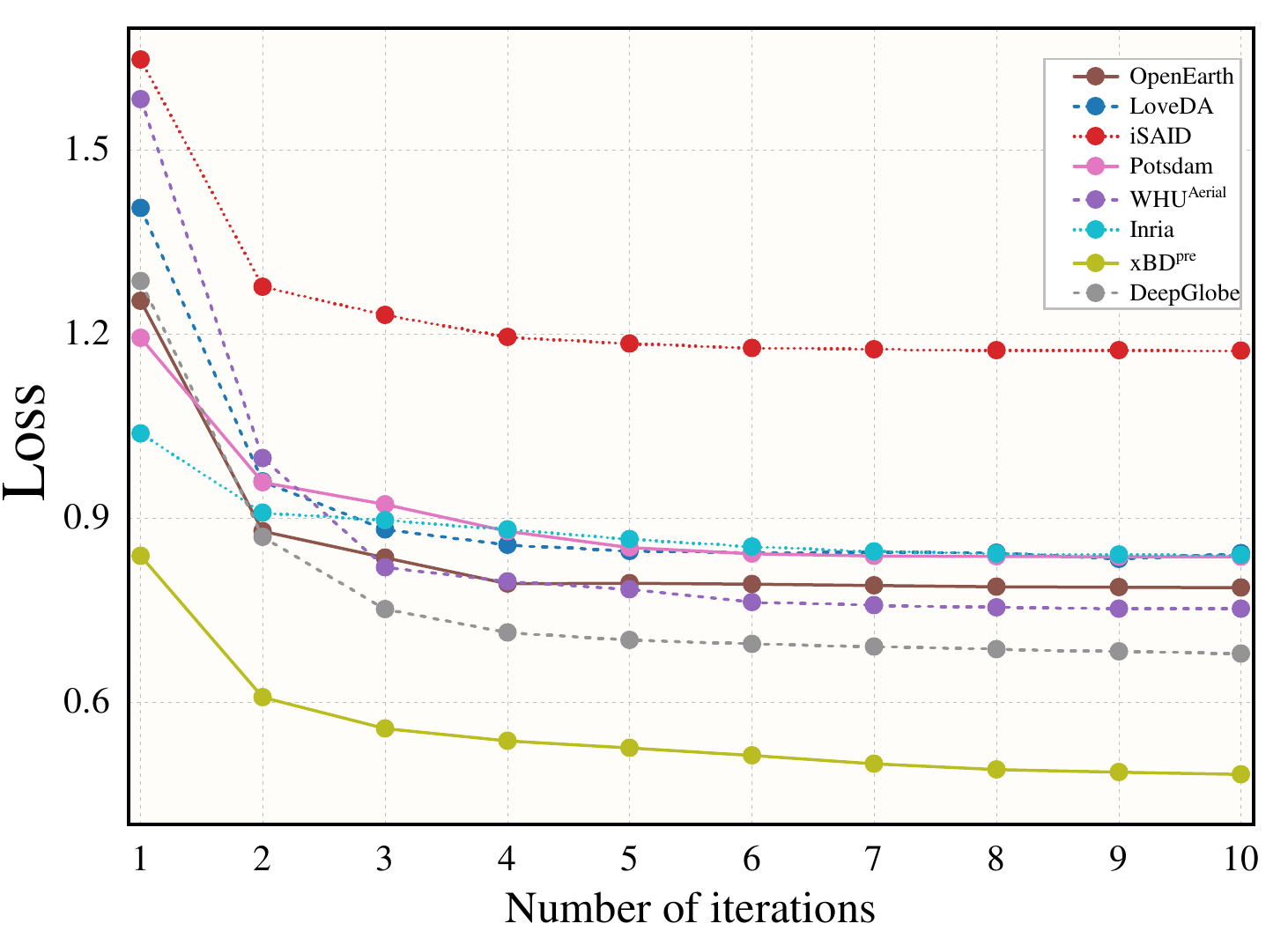}
    \caption{Convergence analysis was conducted on 8 datasets. During the iterative process, the $loss$ rapidly converged and subsequently remained stable.}
    \label{fig:Loss}
\end{figure}

\subsection{Further Analysis}
The convergence behavior of our iterative method is illustrated in~\cref{fig:Loss}. We observe only a modest gap between the initial and final loss values, confirming the effectiveness of our prior prediction-based initialization strategy. The loss decreases rapidly within the first few iterations and stabilizes thereafter, reaching a smooth plateau without noticeable oscillations. This demonstrates that the optimization procedure converges efficiently in practice.

\section{Conclusion}
\label{sec:conclusion}
In this work, we presented \emph{ConInfer}, a training-free context-aware inference framework for open-vocabulary remote sensing segmentation. By explicitly modeling cross-patch semantic dependencies and jointly reasoning over multiple spatial units, \emph{ConInfer} substantially improves segmentation consistency, and consistently surpasses existing training-free and state-of-the-art VLM-based baselines across diverse remote sensing datasets. Our results demonstrate that context-awareness is a critical missing component in training-free OVRSS and provides a new inference-level paradigm beyond independent patch prediction.
Despite these promising results, \emph{ConInfer} currently operates at the patch level, which limits its capability in fine-grained boundary delineation. In future work, we will explore pixel-level contextual modeling to further enrich structural detail, enhance boundary precision, and strengthen generalization across heterogeneous remote sensing domains.
{
    \small
    \bibliographystyle{ieeenat_fullname}
    \bibliography{main}
}

\clearpage
\setcounter{page}{1}
\maketitlesupplementary

\section{Datasets}
\label{Appendix:dataset}
\subsection{Semantic Segmentation}
\begin{itemize}
\item \textbf{OpenEarthMap}~\cite{Openearthmap} includes worldwide satellite and aerial images with a spatial resolution of 0.25-0.5m. It contains 8 foreground classes and one background class. We use its validation set (excluding xBD data) for evaluation. 

\item \textbf{LoveDA}~\cite{LoveDA} is constructed using 0.3m images obtained from the Google Earth platform. It contains both urban and rural areas. It contains 6 foreground classes and one background class. We use its validation set for evaluation. 

\item \textbf{iSAID}~\cite{iSAID} is manily collected from the Google Earth, some are taken by satellite JL-1, the others are taken by satellite GF-2. Its image data is the same as the DOTAv1.0 dataset [69]. It contains 15 foreground classes and one background class. We use its validation set for evaluation, which is cropped to 11,644 images by default (patch size=896, overlap area=384). 

\item \textbf{Potsdam\footnote{https://www.isprs.org/education/benchmarks/UrbanSemLab}} and \textbf{Vaihingen\footnote{https://www.isprs.org/education/benchmarks/UrbanSemLab}} are for urban semantic segmentation used in the 2D Semantic Labeling Contest. Their spatial resolutions are 5cm and 9cm, respectively, and they contain 5 foreground classes and one background class. We use the validation set for evaluation according to MMSeg’s\footnote{https://github.com/open-mmlab/mmsegmentation} setting. 

\item \textbf{UAVid}~\cite{UAVid} consists of 30 video sequences capturing 4K HR images in slanted views. We treat them as images without considering the relationship between frames, and the classes “static car” and “moving car” are converted to “car”. Therefore, it contains 5 foreground classes and one background class. We use its test set for evaluation, which is cropped to 1020 images (patch height=1280, patch width=1080, no overlap). 

\item \textbf{UDD5}~\cite{UDD} is collected by a professional-grade UAV (DJIPhantom 4) at altitudes between 60 and 100m. It contains 4 foreground classes and one background class. We use its validation set for evaluation. 

\item \textbf{VDD}~\cite{vdd} is collected by DJI MAVIC AIR II, including 400 RGB images with 4000*3000 pixel size. All the images are taken at altitudes ranging from 50m to 120m. It contains 6 foreground classes and one background class. We use its test set for evaluation.

\end{itemize}

\subsection{Building extraction}
\begin{itemize}
\item WHUAerial~\cite{WHU} consists of more than 220k independent buildings extracted from aerial images with 0.075m spatial resolution and 450 $km^2$ covering in Christchurch, New Zealand. We use its validation set for evaluation. 

\item WHUSat.II~\cite{WHU} consists of 6 neighboring satellite images covering 860 $km^2$ on East Asia with 0.45m ground resolution. We use its test set (3726 tiles with 8358 buildings) for evaluation. The original images are cropped to 1000 × 1000 without overlap. 

\item Inria~\cite{Inria} covers dissimilar urban settlements, ranging from densely populated areas (e.g., San Francisco’s financial district) to alpine towns (e.g,. Lienz in Austrian Tyrol). It covers 810 $km^2$ with a spatial resolution of 0.3m. We use its test set for evaluation. 

\item xBD~\cite{xBD} covers a diverse set of disasters and geographical locations with over 800k building annotations across over 45k $km^2$ of imagery. Its spatial resolution is 0.8m. We use the pre-disaster satellite data of test set for evaluation.
\end{itemize}

\subsection{Road extraction}
\begin{itemize}
\item CHN6-CUG~\cite{CHN6-CUG} is a large-scale satellite image data set of representative cities in China, collected from Google Earth. It contains 4511 labeled images of 512 × 512 size with a spatial resolution of 0.5m. We use its test set for evaluation. 

\item DeepGlobe\footnote{http://deepglobe.org} covers images captured over Thailand, Indonesia, and India. Its available data cover 362 $km^2$ with a spatial resolution of 5m. The roads are precisely annotated with varying road widths. We use the validation set for evaluation according to the setting in~\cite{Segearth-ov}. 

\item Massachusetts~\cite{Massachusetts} covers a wide variety of urban, suburban, and rural regions and covers an area of over 2,600 $km^2$ with a spatial resolution of 1m. Its labels are generated by rasterizing road centerlines obtained from the OpenStreetMap project, and it uses a line thickness of 7 pixels. We use its test set for evaluation. 

\item SpaceNet~\cite{Spacenet} contains 422 $km^2$ of very high-resolution imagery with a spatial resolution of 0.3m. It covers Las Vegas, Paris, Shanghai, Khartoum and is designed for the SpaceNet challenge. We use the test set for evaluation according to the setting in~\cite{Segearth-ov}.
\end{itemize}

\begin{table*}[ht]
  \centering
  \caption{The prompt class name of the evaluation datasets. {} indicates multiple prompt vocabularies for one class.}
  \label{tab:prompt}
  \begin{tabularx}{\textwidth}{l X}
  \toprule
    \textbf{Dataset} & \textbf{Class Name} \\
    \midrule
    OpenEarthMap & background, \{bareland, barren\}, grass, pavement, road, \{tree, forest\}, \{water, river\}, cropland, \{building, roof, house\} \\
    \addlinespace
    LoveDA & background, roof, road, river, barren, forest, agricultural \\
    \addlinespace
    iSAID & \{background, grass, agriculture, water, forest, road, building, barren, urban\}, boat, industrial cylindrical storage tank, baseball diamond, tennis court, basketball court, ground track field, bridge, large vehicle, small vehicle, helicopter, swimming pool, roundabout, soccer ball field, plane, harbor \\
    \addlinespace
    Potsdam & \{road, parking lot\}, building, low vegetation, tree, car, \{clutter, background\} \\
    \addlinespace
    Vaihingen & impervious surface, building, low vegetation, tree, car, clutter \\
    \addlinespace
    UAVid & background, building, road, car, tree, vegetation, human \\
    \addlinespace
    UDD5 & vegetation, building, road, vehicle, background \\
    \addlinespace
    VDD & \{background, barren\}, exterior, road, grass, car, roof, water \\
    \addlinespace
    WHUAerial & \{forest, vegetation, bareland, river, road, vehicle\}, building \\
    \addlinespace
    WHUSat.II & \{forest, vegetation, bareland, river, route, vehicle\}, building \\
    \addlinespace
    Inria & \{tree, vegetation, bareland, river, paved road, vehicle\}, building \\
    \addlinespace
    xBD & \{background, forest, vegetation, bareland, river, paved road\}, building \\
    \addlinespace
    CHN6-CUG & \{background, forest, vegetation, barren, water, building, vehicle, basketball court, agriculture, urban, rangeland, cropland\}, road \\
    \addlinespace
    DeepGlobe & \{background, forest, vegetation, barren, water, building, vehicle, agriculture, urban, rangeland, cropland\}, road \\
    \addlinespace
    Massachusetts & \{background, forest, vegetation, barren, water, building, vehicle, urban, rangeland, cropland\}, road \\
    \addlinespace
    SpaceNet & \{background, forest, vegetation, barren, water, building, vehicle, agriculture, urban, rangeland, cropland\}, road \\
    \addlinespace
    WBS-SI & \{background, forest, vegetation, barren, agriculture, urban\}, water \\
    \bottomrule
  \end{tabularx}
\end{table*}

\subsection{Flood Detection}
\begin{itemize}
\item WBS-SI\footnote{https://www.kaggle.com/datasets/shirshmall/water-body-segmentation-in-satellite-images} is a satellite image dataset for water body segmentation. It contains 2495 images and we randomly divided 20\% of the data as a test set for evaluation.
\end{itemize}

\section{Class prompt}
We define the prompt for each class as~\cref{tab:prompt}. For single-class extraction tasks, we define the background class as a miscellaneous category independent of the foreground class. This approach is reasonable for clustering and also enhances performance for other methods.

\section{MaskCLIP*~(Baseline)}
Drawing inspiration from the hierarchical architecture of Vision Transformers~\cite{Swin_transformer, Twins, Metaformer}, we divide the ViT into four stages to leverage the transition from local to global attention patterns.
We extract attention maps from the middle layers (2nd, 5th, 8th, 11th) of each stage to form a multi-level attention composite, denoted as $\mathrm{Attn}_{M}$.
This composite is then integrated with the attention map of the final layer, $\mathrm{Attn}$, which is modified according to the SegEarth-OV~\cite{Segearth-ov} setting: its FFN is removed, and attention weights are derived from the sum of query, key, and value self-similarities.
The final fused attention, $\mathrm{Attn}'$, is computed as a weighted sum with a coefficient $\lambda$:
\begin{equation}
    \mathrm{Attn}' = \lambda \cdot \mathrm{Attn} + (1-\lambda) \cdot \mathrm{Attn}_{M}.
    \label{eq:attention_fusion}
\end{equation}
In the experiment, $\lambda$ was set to 0.6. We report in~\cref{tab:Attn1,tab:Attn2} the impact of varying $\lambda$, as well as the effect of removing $\mathrm{Attn}_{M}$ from the framework in~\cref{tab:lambda1,tab:lambda2}.

\begin{table*}[]
  \centering
  \caption{Comparisons of mIoU (448) on different datasets.}
  \label{tab:Attn1}
  \resizebox{\textwidth}{!}{%
    \begin{tabular}{lccccccccc}
    \toprule
    Method & OpenEarthMap & LoveDA & iSAID & Potsdam & Vaihingen & UAVid\_img & UDD5  & VDD   & Average \\
    \midrule
    Baseline (w/o $\mathrm{Attn}_{M}$) & 31.87 & 30.09 & 18.43 & 43.15 & 21.60 & 37.66 & 42.36 & 42.20 & 33.42 \\
    Baseline (w/ $\mathrm{Attn}_{M}$)  & 32.46 & 31.58 & 18.17 & 44.49 & 19.94 & 37.20 & 42.17 & 42.54 & 33.57 \\
    Ours (w/o $\mathrm{Attn}_{M}$)     & 41.35 & 37.89 & 19.29 & 49.20 & 31.36 & 45.92 & 46.54 & 47.95 & 39.94 \\
    Ours (w/ $\mathrm{Attn}_{M}$)      & 41.95 & 39.33 & 20.08 & 49.99 & 31.37 & 46.40 & 46.86 & 50.29 & 40.78 \\
    \bottomrule
    \end{tabular}%
  }
\end{table*}

\begin{table*}[]
  \centering
  \caption{Comparison of Foreground IoU (448) on different datasets.}
  \label{tab:Attn2}
  \resizebox{\textwidth}{!}{%
    \begin{tabular}{lcccccccccc}
    \toprule
    Method & WHU\_Aerial & WHU\_Sat.II & Inria & xBD\_pre & CHN6-CUG & DeepGlobe & Massachusetts & SpaceNet & WBS-SI & Average \\
    \midrule
    Baseline (w/o $\mathrm{Attn}_{M}$) & 45.09 & 17.00 & 40.52 & 34.27 & 27.30 & 14.13 & 8.04  & 19.07 & 48.98 & 28.27 \\
    Baseline (w/ $\mathrm{Attn}_{M}$)  & 45.76 & 15.17 & 37.26 & 31.40 & 29.12 & 14.68 & 8.42  & 19.83 & 52.57 & 28.25 \\
    Ours (w/o $\mathrm{Attn}_{M}$)     & 55.07 & 31.85 & 53.66 & 41.01 & 35.67 & 18.02 & 11.84 & 23.47 & 59.40 & 36.67 \\
    Ours (w/ $\mathrm{Attn}_{M}$)      & 58.54 & 39.35 & 55.65 & 41.34 & 40.00 & 19.85 & 12.16 & 24.02 & 61.38 & 39.14 \\
    \bottomrule
    \end{tabular}%
  }
\end{table*}

\begin{table*}[]
  \centering
  \caption{Performance comparison with different $\lambda$ values across multiple datasets.}
  \label{tab:lambda1}
  \resizebox{\textwidth}{!}{%
    \begin{tabular}{lcccccccc}
    \toprule
    Setting & OpenEarthMap & LoveDA & iSAID & Potsdam & Vaihingen & UAVid\_img & UDD5  & VDD \\
    \midrule
    $\lambda=0.0$ & 41.35 & 37.89 & 19.29 & 49.20 & 31.36 & 45.92 & 46.54 & 47.95 \\
    $\lambda=0.2$ & 41.82 & 38.19 & 19.73 & 49.78 & 31.15 & 46.12 & 46.71 & 48.23 \\
    $\lambda=0.4$ & 42.09 & 38.90 & 19.97 & 49.93 & 30.62 & 46.14 & 46.82 & 49.31 \\
    $\lambda=0.6$ & 41.95 & 39.33 & 20.08 & 49.99 & 29.73 & 45.60 & 46.86 & 50.29 \\
    $\lambda=0.8$ & 40.06 & 39.61 & 19.86 & 49.83 & 27.74 & 44.59 & 46.93 & 48.43 \\
    $\lambda=1.0$ & 39.92 & 39.31 & 19.35 & 49.35 & 25.92 & 43.43 & 44.58 & 46.39 \\
    \bottomrule
    \end{tabular}%
  }
\end{table*}

\begin{table*}[]
  \centering
  \caption{Performance comparison on multiple datasets with varying $\lambda$.}
  \label{tab:lambda2}
  \resizebox{\textwidth}{!}{%
    \begin{tabular}{lccccccccc}
    \toprule
    Setting & WHU\_Aerial & WHU\_Sat.II & Inria & xBD\_pre & CHN6-CUG & DeepGlobe & Massachusetts & SpaceNet & WBS-SI \\
    \midrule
    $\lambda=0.0$ & 55.07 & 31.85 & 53.66 & 41.01 & 35.67 & 18.02 & 11.84 & 23.47 & 59.40 \\
    $\lambda=0.2$ & 56.23 & 34.49 & 54.58 & 41.62 & 36.28 & 18.63 & 12.16 & 23.67 & 59.84 \\
    $\lambda=0.4$ & 57.47 & 36.92 & 55.45 & 41.86 & 36.33 & 19.37 & 12.16 & 23.81 & 60.65 \\
    $\lambda=0.6$ & 58.54 & 39.35 & 55.65 & 41.82 & 40.00 & 19.85 & 12.16 & 24.02 & 61.38 \\
    $\lambda=0.8$ & 58.97 & 41.84 & 55.36 & 41.13 & 37.00 & 20.07 & 12.13 & 24.14 & 61.79 \\
    $\lambda=1.0$ & 57.41 & 41.64 & 54.22 & 39.28 & 33.73 & 19.89 & 12.14 & 24.34 & 62.25 \\
    \bottomrule
    \end{tabular}%
  }
\end{table*}

\section{Qualitative Results.}

\begin{figure*}[]
    \centering
    \includegraphics[width=1.0\linewidth]{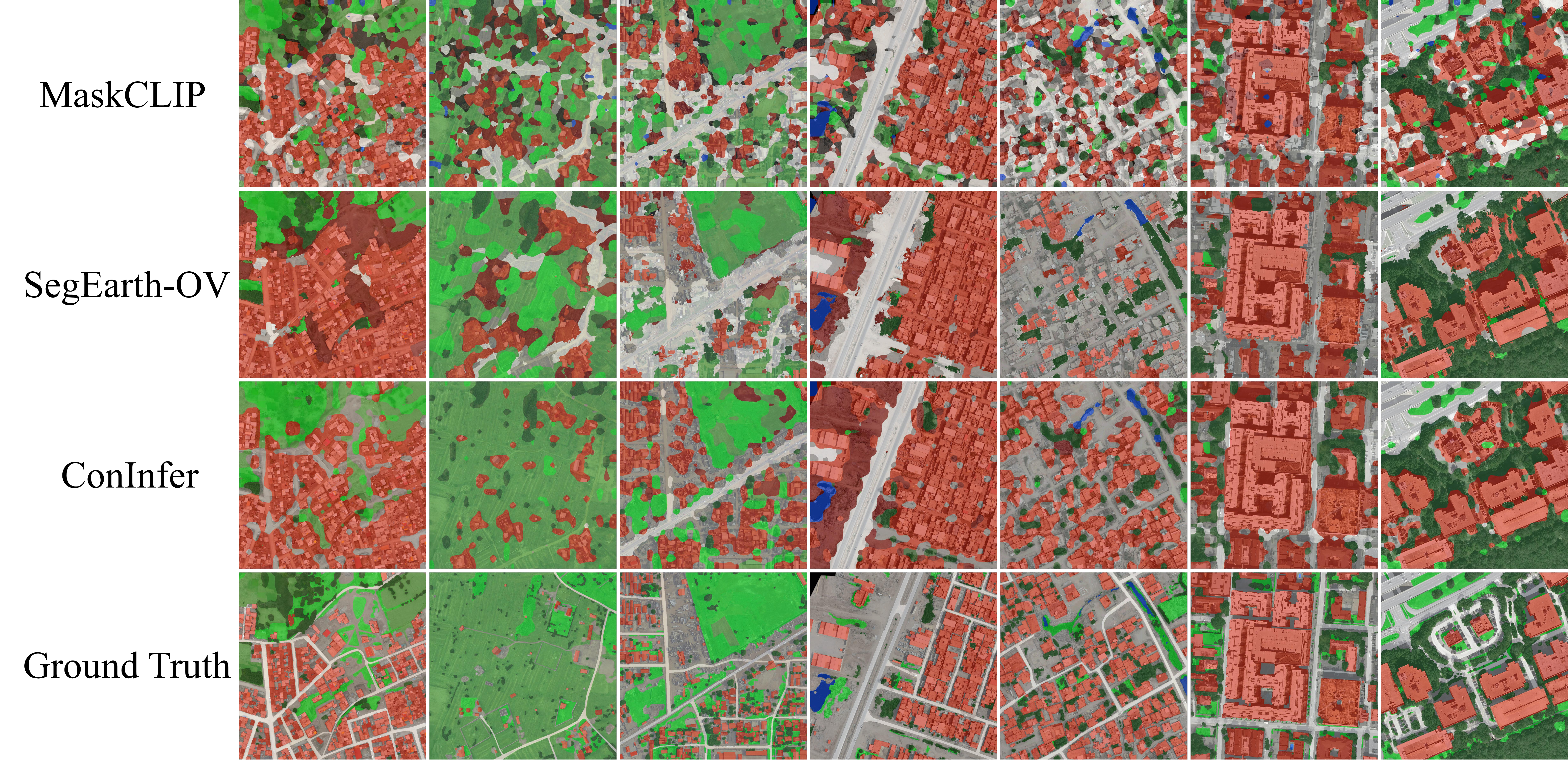}
    \caption{Qualitative comparison of different OVSS methods on the OpenEarthMap (land cover mapping) datasets.}
    \label{fig:Qualitative1}
\end{figure*}

\begin{figure*}[]
    \centering
    \includegraphics[width=1.0\linewidth]{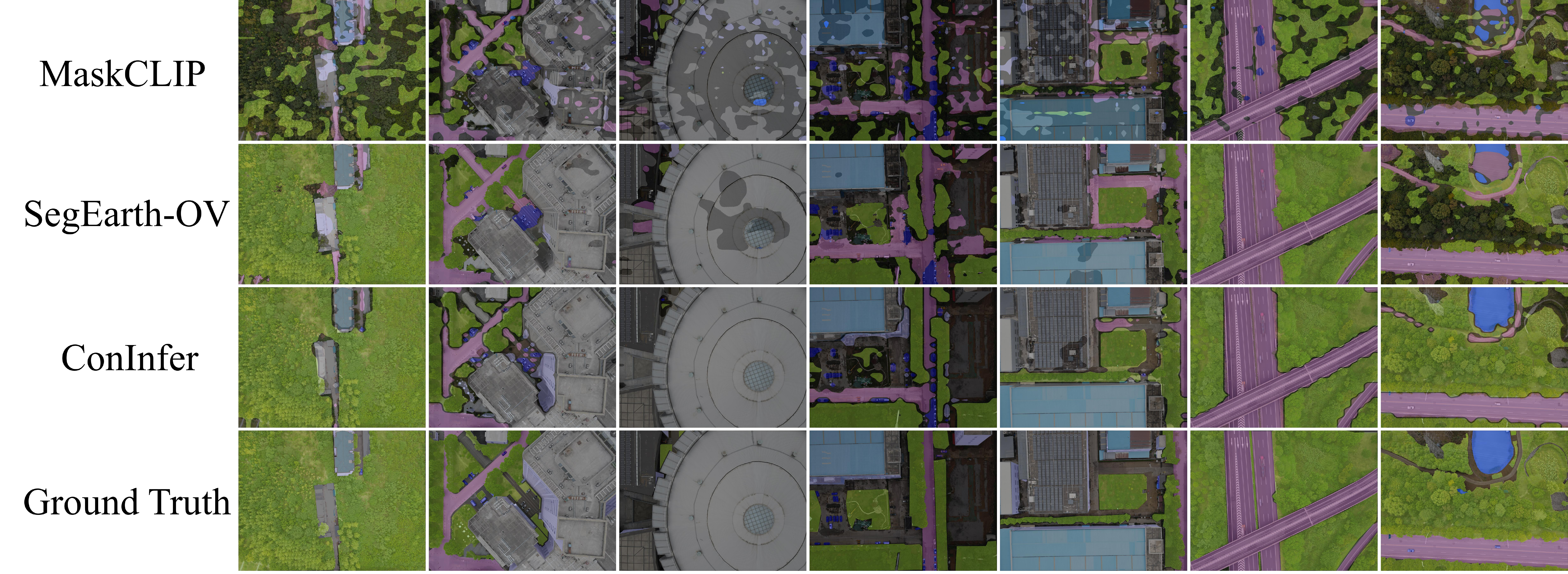}
    \caption{Qualitative comparison of different OVSS methods on the VDD (UAV aerial scene) datasets.}
    \label{fig:Qualitative2}
\end{figure*}

\begin{figure*}[]
    \centering
    \includegraphics[width=1.0\linewidth]{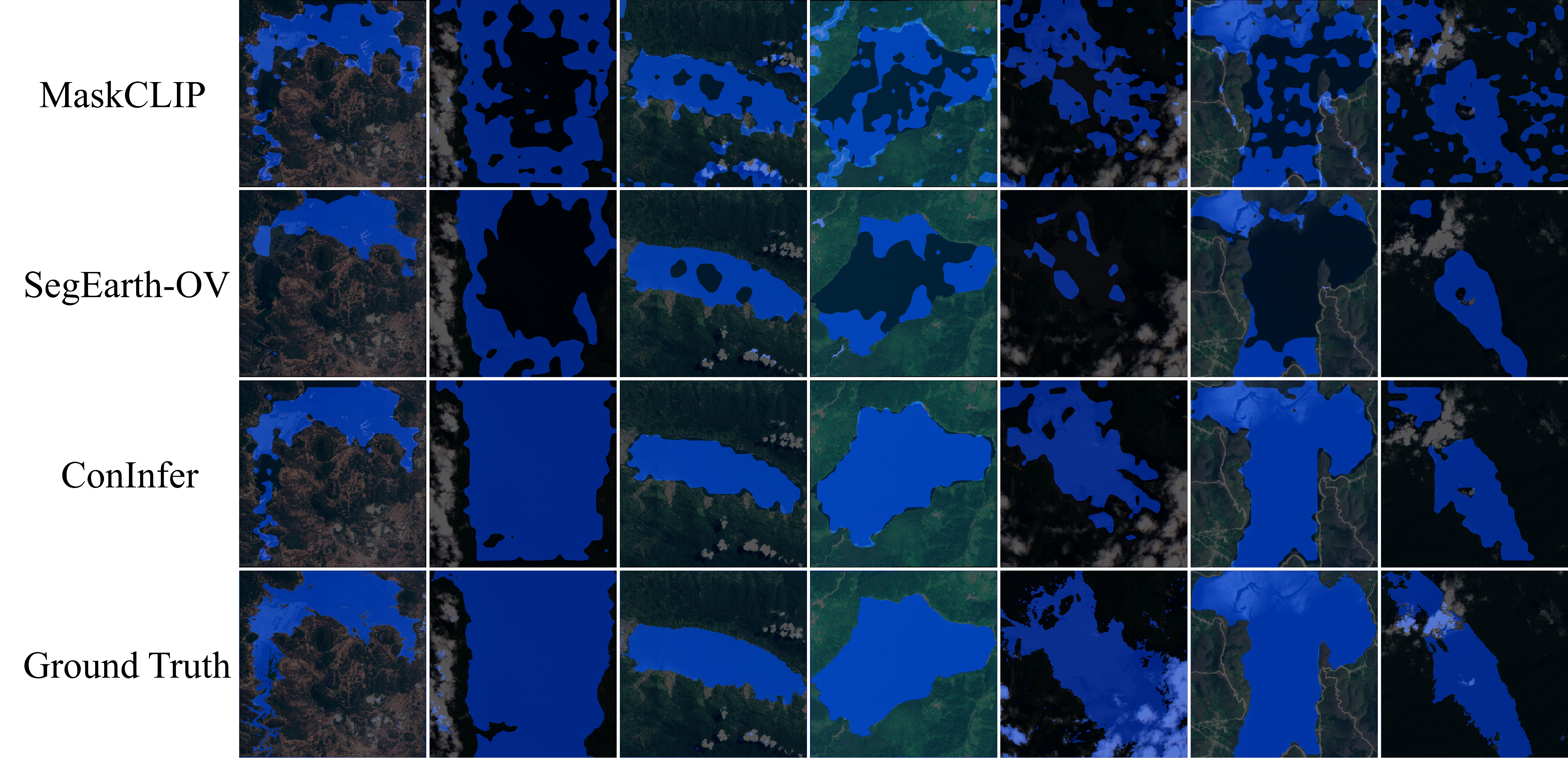}
    \caption{Qualitative comparison of different OVSS methods on the WBI-SI (water body segmentation) datasets.}
    \label{fig:Qualitative3}
\end{figure*}

\end{document}